\def\year{2021}\relax
\documentclass[letterpaper]{article} % DO NOT CHANGE THIS
\usepackage{aaai21}  % DO NOT CHANGE THIS
\usepackage{times}  % DO NOT CHANGE THIS
\usepackage{helvet} % DO NOT CHANGE THIS
\usepackage{courier}  % DO NOT CHANGE THIS
\usepackage[hyphens]{url}  % DO NOT CHANGE THIS
\usepackage{graphicx} % DO NOT CHANGE THIS
\usepackage{natbib}  % DO NOT CHANGE THIS AND DO NOT ADD ANY OPTIONS TO IT
\usepackage{caption} % DO NOT CHANGE THIS AND DO NOT ADD ANY OPTIONS TO IT
\usepackage{multirow}
\usepackage{booktabs}
\usepackage{amssymb}
\title{Question-Interlocutor Scope Realized Graph Modeling over Key Utterances for Dialogue Reading Comprehension}
\author{
    Jiangnan Li\textsuperscript{\rm 1,2}, 
    Mo Yu\textsuperscript{\rm 3}, 
    Fandong Meng\textsuperscript{\rm 3}, 
    Zheng Lin\textsuperscript{\rm 1,2}, 
    Peng Fu\textsuperscript{\rm 1}, 
    Weiping Wang\textsuperscript{\rm 1}, 
    Jie Zhou\textsuperscript{\rm 3}
    \\
}
\affiliations{
        \textsuperscript{\rm 1}Institute of Information Engineering, Chinese Academy of Sciences, Beijing, China \\
    \textsuperscript{\rm 2}School of Cyber Security, University of Chinese Academy of Sciences, Beijing, China \\
    \textsuperscript{\rm 3}Pattern Recognition Center, WeChat AI, Tencent Inc, China\\
    \{lijiangnan, linzheng, fupeng, wangweiping\}@iie.ac.cn, \{moyumyu,fandongmeng,withtomzhou\}@tencent.com
}
\begin{document}

\maketitle

\begin{abstract}
In this work, we focus on dialogue reading comprehension (DRC), a task extracting answer spans for questions from dialogues. Dialogue context modeling in DRC is tricky due to complex speaker information and noisy dialogue context. To solve the two problems, previous research proposes two self-supervised tasks respectively: guessing who a randomly masked speaker is according to the dialogue and predicting which utterance in the dialogue contains the answer. Although these tasks are effective, there are still urging problems: (1) randomly masking speakers regardless of the question cannot map the speaker mentioned in the question to the corresponding speaker in the dialogue, and ignores the speaker-centric nature of utterances. This leads to wrong answer extraction from utterances in unrelated interlocutors' scopes; (2) the single utterance prediction, preferring utterances similar to the question, is limited in finding answer-contained utterances not similar to the question. 
%cannot tell the model which utterances belong to a specific interlocutor, which may lead to wrong answer extraction from the utterance scope of unrelated interlocutors. 
To alleviate these problems, we first propose a new key utterances extracting method. It performs prediction on the unit formed by several contiguous utterances, which can realize more answer-contained utterances. Based on utterances in the extracted units, we then propose Question-Interlocutor Scope Realized Graph (QuISG) modeling. As a graph constructed on the text of utterances, QuISG additionally involves the question and question-mentioning speaker names as nodes. To realize interlocutor scopes, speakers in the dialogue are connected with the words in their corresponding utterances. Experiments on the benchmarks show that our method can achieve better and competitive results against previous works.
% \footnote{The code will be publicly available once accepted. }

% and indicate the efficacy of our method. 
\end{abstract}

\section{Introduction}

%People process various types of text in the daily life. 
Beyond the formal forms of text, dialogues are one of the most frequently used media that people communicate with others to informally deliver their emotions~\cite{MELD}, opinions~\cite{StanceConv}, and intentions~\cite{CoGAT}.
%People receive and process various types of textual information in their daily lives beyond the formal text forms. 
%Among them, texts in the dialogue form plays a central role. People communicate with others with dialogues in most of their times, to understand the messages from outside, deliver their own opinions and intentions.
Moreover, dialogues are also a crucial type of information carriers in literature, such as novels and movies~\cite{NarrativeQA}, for people to understand the characters and plots~\cite{TVShowguess} in their reading and entertainment behaviors. Therefore, how to comprehend dialogues is a key step for machines to act like human. 

\begin{figure}
    \centering
    \scalebox{0.4}{\includegraphics{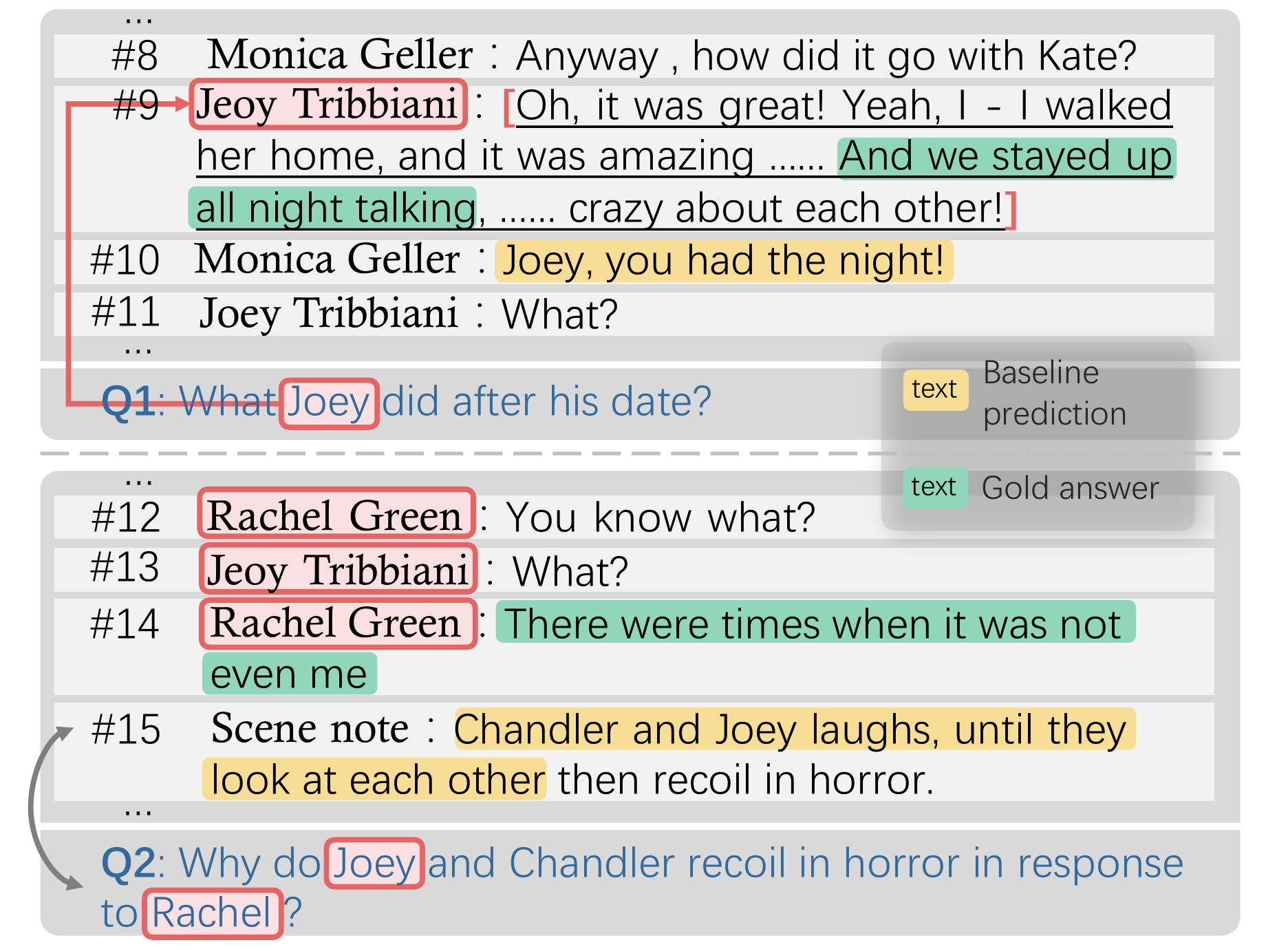}
    }
    \caption{Two questions with related dialogue clips that the baseline SelfSuper~\citep{SelfSuper} fails. Utter. \#9 is too long, so we omit some parts of the utterance. }
    \label{fig:case}
\end{figure}

Despite the importance and value of dialogues, reading comprehension over dialogues (DRC) lags behind those over formal text, e.g., news articles and Wikipedia text.\footnote{Note there is a direction of conversational question answering~\cite{CoQA,QuAC} differing from the DRC task here. For the former, QA are formed as a dialogue, and the model is required to understand \textbf{Wikipedia articles} to derive the answers.}
There are several reasons for the challenges in dialogue reading comprehension.
(1) As shown in Fig.~\ref{fig:case}, utterances in dialogues involve mostly informal \emph{oral language}.
(2) An utterance itself is usually short and incomplete, and therefore understanding it highly depends on its \emph{dialogue context}.
Finally, (3) dialogue texts can be fluctuated. For example, people make a slip of language sometimes, and may evoke emotional expression. 
%\mo{There is a problem about the interlocutor scope: the ablation study shows that taking off scope-related edges does not affect the performance that much as w/o Q and w/o KeyUtter do. }Utter. \#9 in the Figure gives such an example., e.g., utter. \#10 in Fig.~\ref{fig:case}
The aforementioned challenges in (1) and (3) can be alleviated with pretrained models with domain-specific training data and enhancement of robustness.
However, the challenge (2) is a major scientific problem in DRC. 

In previous works, \citet{SelfSuper} (abbreviated as SelfSuper) point out that \emph{dialogue context} modeling in DRC faces two challenges: complex speaker information and noisy question-unrelated context. For speaker modeling, SelfSuper design a self-supervised task guessing who a randomly masked speaker is according to the dialogue context (e.g., masking ``\texttt{Monica Geller}'' of \#10). To reduce noise, SelfSuper design another task to predict whether an utterance contains the answer. Although decent performance can be achieved, several urging problems still exist. 

Firstly, speaker guessing does not aware the speaker information in questions and the interlocutor scope. As randomly masking is independent to the question, it cannot tell which speaker in the dialogue is related to the speaker mentioned in the question, e.g., \texttt{Joey Tribbiani} to \texttt{Joey} in Q1 of Fig.~\ref{fig:case}. As for the interlocutor scope, we define it as the utterances said by the corresponding speaker. We point that utterances have a speaker-centric nature: First, each utterance has target listeners. For example, in Utter. \#10 of Fig.~\ref{fig:case}, it requires to understand that \texttt{Joey} is a listener, so ``\texttt{you had the night}'' is making fun of \texttt{Joey} from \texttt{Monica}'s scope. Second, an utterance reflects the message of experience of its speaker. For example, to answer Q1 in Fig.~\ref{fig:case}, it requires to understand ``\texttt{stayed up all night talking}'' is the experience appearing in \texttt{Joey}'s scope. Due to ignoring the question-mentioned interlocutor and its scope, SelfSuper provides a wrong answer. 
%Finally, people frequently use the other interlocutors' names to refer the context they hope to respond to or address.\mo{can we make a different example to replace the bottom, something like \#10 but with the mention as a reference to the previous utterance instead of the second person usage.}  to what he described earlier. the speech act or \

Secondly, answer-contained utterance (denoted as key utterance by SelfSuper) prediction prefers utterances similar to the question, failing to find key utterances not similar to the question. The reason for this is that answers are likely to appear in utterances similar to the question. For example, about 77\% questions has answers in top-5 utterances similar to the question according to SimCSE~\cite{SimCSE} in the dev set of FriendsQA~\cite{FriendsQA}. Furthermore, the utterances extracted by the key utterance prediction have over 82\% overlaps with the top-5 utterances. Therefore, there are considerable key utterances have been ignored, leading to overrated attention to similar utterances, e.g., Q2 in Fig.~\ref{fig:case}. In fact, answer-contained utterances are likely to appear as the similar utterances or near the similar utterances because contiguous utterances in local context tend to be in a topic relevant to the question. However, the single utterance prediction cannot realize this observation. 

To settle aforementioned problems, so that more answer-contained utterances can be found and answering process realizes the question and interlocutor scopes, we propose a new pipeline framework for DRC. We first propose a new key utterances extracting method. The method slides a window through the dialogue, where contiguous utterances in the window are regarded as a unit. The prediction is made on these units. Once a unit containing the answer, all utterances in it are selected as key utterances. Based on these extracted utterances, we then propose Question-Interlocutor Scope Realized Graph (QuISG) modeling. Instead of treating utterances as a plain sequence, QuISG constructs a graph structure over the contextualized text embeddings. The question and speaker names mentioned in the question are explicitly present in QuISG as nodes. The question-mentioning speaker then connects with its corresponding speaker in the dialogue. Furthermore, to remind the model of interlocutor scopes, QuISG connects every speaker node in the dialogue with words from the speaker's scope. We verify our model on two representative DRC benchmarks, FriendsQA and Molweni~\cite{Molweni}. Our model achieves better and competitive performance against baselines on both benchmarks, and further experiments indicate the efficacy of our proposed method. 

\section{Related Work}

\paragraph{Dialogue Reading Comprehension.}
Unlike traditional Machine Reading Comprehension~\cite{SQuAD}, Dialogue Reading Comprehension (DRC) aims to answer a question according to the given dialogue. There are several related but different types of conversational question answering: CoQA~\cite{CoQA} conversationally asks questions after reading Wikipedia articles. QuAC~\cite{QuAC} forms a dialogue of QA between a student and a teacher about Wikipedia articles. DREAM~\cite{DREAM} tries to answer multi-choice questions over dialogues of English exams. To understand characteristics of speakers, \citet{TVShowguess} propose TVShowGuess in a multi-choice style to predict unknown speakers in dialogues. Conversely, we focus on DRC extracting answer spans from a dialogue for an independent question~\cite{FriendsQA}. For our focused DRC, as dialogues are a kind of special text, \citet{ULMUOP} propose several pretrained and downstream tasks based on utterances in dialogues. To consider coreference and relationship of speakers, \citet{KnowledgeGraph} introduce the two types of knowledge from other dialogue-related tasks. Besides, \citet{DADGraph,speakerAware} model the knowledge of discourse structure for dialogues. To model the complex speaker information and noisy dialogue context, \citet{SelfSuper} propose two self supervised tasks, i.e., masked-speaker guessing and key utterance prediction. However, existing work ignores explicitly modeling the question and speaker scopes and suffers from low key-utterance coverage. 

\begin{figure*}
    \centering
    \scalebox{0.43}{\includegraphics{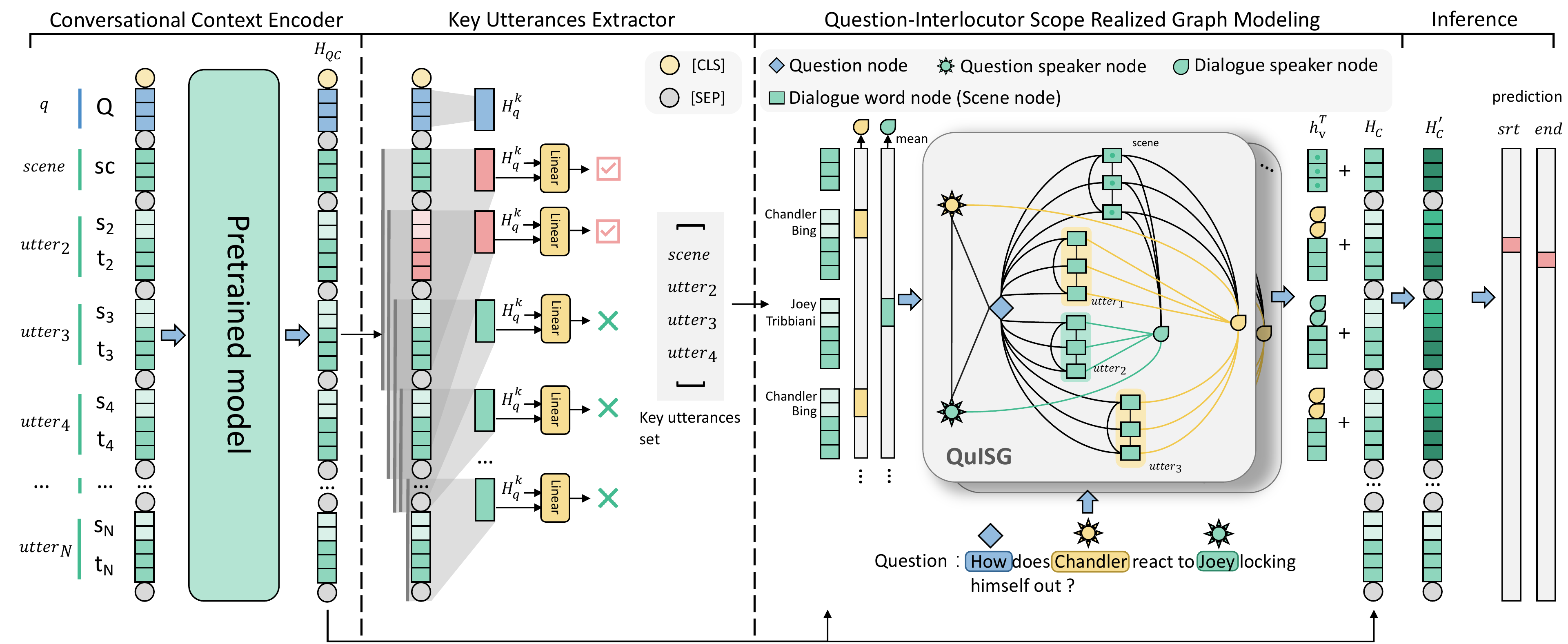}}
    \caption{The overall framework of our proposed model. We first encode the dialogue and the question by pretrained models. Key utterances extractor takes contiguous utterances as a unit to extract key utterances. Based on extracted key utterances, question-interlocutor scope realized graph is constructed. }
    \label{main}
\end{figure*}

\paragraph{Dialogue Modeling with Graph Representations. }
In many QA tasks~\cite{HotpotQA,CommonsenseQA}, graphs are the main carrier for reasoning~\cite{DFGN,HiGraph,QAGNN}. As for dialogue understanding, graphs are still a hotspot for various purposes. In dialogue emotion recognition, graphs are constructed to consider the interactions between different parties of speakers~\citep{DialogueGCN,posEmb,DAG}. In dialogue act classification, graphs model the cross-utterances and cross-tasks information~\cite{CoGAT}. In dialogue semantic modeling, \citet{semanticRepresentation} extend AMR~\cite{AMR} to construct graphs for dialogues. As for our focused DRC, graphs are constructed for knowledge propagation between utterances by works~\cite{KnowledgeGraph,DADGraph,speakerAware} mentioned above. 

%However, they ignore the interlocutor scope and the question-related information. \mo{I was trying to argue that they use graphs to achieve different purposes for DRC.Or there is not much graph works for DRC, but in general RC?}
% (1) retrieval of evidence in open-domain QA, \cite{danqi,}
% (2) incorporation of commonsense knowledge~\cite{findsomeref,yourownwork}
% (3) handle long inputs~\cite{shuohang}

\section{Framework}

\subsection{Task Definition}

Given a dialogue consisting of \textit{N} utterances: $\mathcal{D}$ = $[utter_{1}$ $, utter_{2}, ..., utter_{N}]$, the task aims to extract the answer span $a$ for a question $q=[qw_1, qw_2, ..., qw_{L_q}]$ from $\mathcal{D}$, where $qw_i$ is the i-\textit{th} word in $q$ and $L_q$ is the length of $q$. In $\mathcal{D}$, each utterance $utter_{i}=\{\mathrm{speaker}: s_i, \mathrm{text}: t_i\}$ contains its corresponding speaker (e.g., $s_i=$``\texttt{Chandler Bing}") and text content $t_i=[tw_1, tw_2, ...,tw_{L_i}]$, where $tw_j$ the j-\textit{th} word in $t_i$ and $L_i$ is the length of $t_i$. For some unanswerable questions, there is no answer span to be found in $\mathcal{D}$. Under such a circumstance, $a$ is assigned to be null. 

\subsection{Conversational Context Encoder}

Dialogue is a process whose utterances are not isolated but contextually interact with each other. Therefore, encoding words in an utterance requires considering contextual information from other utterances. To better encode words contextually using pretrained models (PTM), following previous work \citep{SelfSuper}, we chronologically concatenate utterances in the same conversation to form a text sequence: $\mathcal{C}$ = ``$s_1$: $t_1$ $\mathrm{[SEP]}$ ... $\mathrm{[SEP]}$ $s_N$: $t_N$'', where [SEP] is a special token of PTM for separation. Holding the conversational context $\mathcal{C}$, PTM can deeply encode $\mathcal{C}$ with the question $q$ to make it question-aware by concatenating them as $QC$ = ``$\mathrm{[CLS]}\ q\ \mathrm{[SEP]}\ \mathcal{C}\ \mathrm{[SEP]}$'' (it is okay that $\mathcal{C}$ goes first), where [CLS] is also a special token for classification. 

Following \citet{SelfSuper}, we utilize the ELECTRA discriminator to encode the sequence $QC$:
\begin{equation}
    H_{QC}=\mathrm{ELECTRA}(QC),
\end{equation}
where $H_{QC}\in \mathbb{R}^{L_{QC}\times d_{h}}$, $L_{QC}$ is the length of $QC$, and $d_{h}$ is the hidden size of PLM. $H_{QC}$ can be split into $H_{Q}\in \mathbb{R}^{L_q\times d_{h}}$ and $H_{C}\in \mathbb{R}^{L_C}$ according the position of [SEP] between $q$ and $C$, where $L_C$ is the length of $\mathcal{C}$. 

\subsection{Key Utterances Extractor}

Treating every single utterance as a unit to pair with the question prefers utterances similar to the question. However, the utterance containing the answer is not always in the case, where it can appear near the similar utterance within several steps due to the topic relevance. Therefore, we apply a window along the dialogue. Utterances in the window are treated as a unit so that the similar utterance and the answer-contained utterance can co-occur and more answer-contained utterances can be realized. 

\subsubsection{Training the Extractor.}

With the window whose size is $m$, [$utter_{i}$, $utter_{i+1}$, ..., $utter_{i+m}$] is grouped. Mapping the start ($st_i$) and end ($ed_i$) position of the unit in $\mathcal{C}$, the representation of the unit can be computed by: 
\begin{equation}
    H_{u_i}^{k}=\mathrm{Maxpooling}(H_{C}[st_{i}: ed_{i}]). 
\end{equation}

Similarly, the representation of the question is computed by $H_{q}^{k}=\mathrm{Maxpooling}(H_{Q})$. The correlation score of the unit and the question is then computed by:
\begin{equation}
    y_{i}=\mathrm{sigmoid}(\mathrm{Linear}(H_{u_i}^{k}||H_{q}^{k})), 
\end{equation}
where $\mathrm{Linear}(\cdot)$ is a linear unit mapping the dimension from $\mathbb{R}^{2d_{h}}$ to $\mathbb{R}$ and $||$ is the concatenating operation. For the unit, if any utterances in it contains the answer, the label $y_{i}^{k}$ of this unit is set to 1, otherwise 0. Therefore, the training objective of the key utterances extractor on the dialogue $\mathcal{D}$ is:
\begin{equation}
    \mathcal{J}_{k}=-\sum_{i=1}^{N-m}{[(1-y_{i}^{k})log(1-y_{i})+y_{i}^{k}log(y_{i})]}. 
\end{equation}

\subsubsection{Extracting key utterances.}

The trained model predicts whether a unit is correlated to the question. If $y_{i}>0.5$, the unit is regarded as a question-related unit and utterances inside are all regarded as \textbf{key utterances}. To avoid involving too many utterances as key utterances, we further rank all the units by $y_{i}$ and pick up top-$k$ units whose $y_{i}>0.5$. For a question $q$, we keep a key utterance set $key=(\cdot)$ to store the extracted key utterances. 

Specifically, when the $i$-th unit is among the top-$k$ units and holds $y_{i}>0.5$, [$utter_i$, ..., $utter_{i+m}$] are all considered to be added into $key$. If $utter_i$ does not exist in $key$, then $key$.\texttt{add}($utter_i$) is triggered, otherwise skipped. After processing all units satisfying the conditions, $key$ sorts key utterances in it by \texttt{sort}($key$, 1$\rightarrow$N), where 1$\rightarrow$N denotes chronological order. 

We obverse that, in most cases, key utterances in $key$ are contiguous utterances. When k=3 and m=2, the set is ordered as ($utter_{i-m}$, ..., $utter_{i}$, ..., $utter_{i+m}$), where $utter_{i}$ is usually the similar utterance. 
% with the similar utterance centered (e.g., , where $utter_{i}$ is a most utterance to $q$). The answer-contained utterance can be similar utterance, and it can also appear in the local context of the most similar utterance. 

\subsection{Question-Interlocutor Scope Realized Graph Modeling}

To guide models to further realize the question, speakers in the question, and the scope of the corresponding speakers in $\mathcal{D}$, we construct a Question-Interlocutor Scope Realized Graph (QuISG) based on key utterances in $key$. QuISG can be formulated as $\mathcal{G}$ = ($\mathcal{V}$, $\mathcal{A}$), where $\mathcal{V}$ denotes the set of nodes involved in $\mathcal{G}$ and $\mathcal{A}$ denotes the adjacent matrix to indicate edges. After the construction of QuISG, we utilize a node type realized graph attention network to process it. We elaborate the graph modeling below. 

\subsubsection{Nodes.}
QuISG is directly constructed on the words of the question and key utterances. Therefore, we define several types of node based on words. 

\textit{Question Node}: Question node denotes the questioning word (e.g., ``what'', ``for what reason'') of the question. The node representation is initialized by the meanpooling on the representations of the question word:  $\mathrm{v.rep}$=$\mathrm{mean}(H_{Q}[$for, what, reason$])$. We denote this type of node as $\mathrm{v.type}$=\texttt{qw}. 

\textit{Question Speaker Node}: Considering speakers in the question can help models to realize which speakers and their interactions are focused by the question. Question speaker node is derived from the speaker name recognized from the question. We use stanza \footnote{\url{https://github.com/stanfordnlp/stanza}} performing NER to recognize person names (e.g. ``ross'') in the question and pick up those names appearing in the dialogue as interlocutors. The representation of node is initialized as $\mathrm{v.rep}$=$H_Q[$ross$]$ and the type is marked as $\mathrm{v.type}$=\texttt{qs}. Additionally, if a question contains no speaker name or the picked name does not belong to interlocutors in the dialogue, no question speaker node will be involved. 

\textit{Dialogue Speaker Node}: Speakers appeared in the dialogue are crucial for dialogue modeling. We consider speakers of the key utterances to construct dialogue speaker nodes. As the speaker in the dialogue is identified by its full name (e.g., ``Ross Gellar''), we compute the initial node representation by meanpooling on the full name and all utterances of the speaker will provide its speaker name for the computation: $\mathrm{v.rep}$=$\mathrm{mean}(H_C[$Ross$_{1}$, Gellar$_{1}$, ..., Ross$_{x}$, Gellar$_{x}])$, where $x$ is the number of utterance whose speaker name is ``Ross Gellar''. This type is marked as $\mathrm{v.type}$=\texttt{ds}. 

\textit{Dialogue Word Node}: As the main body where extracting answer is performed, words from utterances are positioned in the graph as dialogue word nodes. We take all words in the key utterances to form the group of dialogue word node. The representation is initialized from the corresponding item of $H_C$. This type is denoted as $\mathrm{v.type}$=\texttt{dw}. 

\textit{Scene Node}: In some datasets, there is a kind of special utterance that appears at the beginning of a dialogue and gives a brief description of where the dialogue happens or what it is about. If it is selected as a key utterance, we set all words in it as scene nodes. Although we define this type of node, it still acts as a word node with $\mathrm{v.type}$=\texttt{dw}. The only difference is how to connect with dialogue speaker nodes, which is stated in the next subsection.

\subsubsection{Edges.}
We set edges between nodes to connect nodes from the question and key utterances and initialize the adjacent matrix as $\mathcal{A}=\mathrm{O}$. 

For the word node $\mathrm{v}_{x} \in utter_{i}$, we connect it with other word nodes $\mathrm{v}_{y} \in utter_{i}$ ($x-k_{w}$ $\leq$ $y$ $\leq$ $x+k_{w}$) within a window whose size is $k_{w}$, i.e., $\mathcal{A}[\mathrm{v}_x,\mathrm{v}_y]=1$. For word nodes in other utterances (e.g., $\mathrm{v}_z\in utter_{i+1}$), there is no edge between $\mathrm{v}_x$ and $\mathrm{v}_z$. To remind the model of the scope of speakers, we connect every word node with the speaker node $\mathrm{v}_{s_i}$ it belongs to, i.e., $\mathcal{A}[\mathrm{v}_x,\mathrm{v}_{s_i}]=1$ and $\mathcal{A}[\mathrm{v}_{s_i},\mathrm{v}_x]=1$. To realize the question, we connect all word nodes with the question node $\mathrm{v}_q$, i.e., $\mathcal{A}[\mathrm{v}_x,\mathrm{v}_{q}]=1$ and $\mathcal{A}[\mathrm{v}_q,\mathrm{v}_{x}]=1$. 

For the speakers mentioned in the question, we fully connect their corresponding question speaker nodes to model the interactions between these speakers, e.g., $\mathcal{A}[\mathrm{v}_{qs_m},\mathrm{v}_{qs_n}]=1$ and $\mathcal{A}[\mathrm{v}_{qs_n},\mathrm{v}_{qs_m}]=1$. To remind the model which speaker in dialogue is related, we further connect the question speaker node $\mathrm{v}_{qs_m}$ with its corresponding speaker node $\mathrm{v}_{s_i}$, i.e., $\mathcal{A}[\mathrm{v}_{qs_m},\mathrm{v}_{s_i}]=1$ and $\mathcal{A}[\mathrm{v}_{s_i},\mathrm{v}_{qs_m}]=1$. Furthermore, to realize the question, question speaker nodes should connect with the question node, e.g., $\mathcal{A}[\mathrm{v}_{qs_m},\mathrm{v}_{q}]=1$ and $\mathcal{A}[\mathrm{v}_{q},\mathrm{v}_{qs_m}]=1$. 

If the scene description is selected as a key utterance, it will be regarded as an utterance without speaker identification. We treat a scene node as a word node and follow the same edge construction as word nodes. As the scene description is likely to tell things about speakers, we utilize stanza to recognize speakers and connect all scene nodes with the dialogue speaker nodes the recognized speakers related to. 

For every node in QuISG, we additionally add a self-connected edge, i.e., $\mathcal{A}[\mathrm{v},\mathrm{v}]=1$. By realizing these connections mentioned above, items in $\mathcal{A}$ are assigned with 0 or 1 for edge existence. 

\subsubsection{Node Type Dependent Graph Attention Network. }
Node Type Dependent Graph Attention Network (GAT) is a $T$-layer stack of graph attention blocks \citep{GAT}. The input of GAT is a QuISG and GAT propagates and aggregates messages between nodes through edges. We initial the graph representation by $h_{v}^{0}=\mathrm{v.rep}$. We then introduce how the $t$-$th$ layer gathers messages for nodes from the neighbors. 

A graph attention block mainly performs multi-head attention computing, in which each head has the similar operation. We exemplify the attention computing by one head. To measure how important the node $\mathrm{v}_n$ to the node $\mathrm{v}_m$, the node type realized attentive weight is computed by:
% \begin{scriptsize}
% \begin{equation}
%     \alpha_{mn}=\frac{\mathrm{exp}\left(\mathrm{LReLU}\left(\mathrm{a}^{T}[\mathrm{W}_{q}[h_{\mathrm{v}_m}^{t-1}||r_{{\mathrm{v}_m}.\mathrm{type}}]||\mathrm{W}_{k}[h_{\mathrm{v}_n}^{t-1}||r_{{\mathrm{v}_n}.\mathrm{type}}]]\right)\right)}{\sum_{\mathrm{v}_o\in\mathcal{N}_{\mathrm{v}_m}}{\mathrm{exp}\left(\mathrm{LReLU}\left(\mathrm{a}^{T}[\mathrm{W}_{q}[h_{\mathrm{v}_m}^{t-1}||r_{{\mathrm{v}_m}.\mathrm{type}}]||\mathrm{W}_{k}[h_{\mathrm{v}_o}^{t-1}||r_{{\mathrm{v}_o}.\mathrm{type}}]]\right)\right)}}
% \end{equation}
% \end{scriptsize}
% \begin{small}
\begin{eqnarray}
\alpha_{mn}=\frac{\mathrm{exp}\left(\mathrm{LeakyReLU}\left(\mathrm{c}_{mn}\right)\right)}{\sum_{\mathrm{v}_o\in\mathcal{N}_{\mathrm{v}_m}}{\mathrm{exp}\left(\mathrm{LeakyReLU}\left(\mathrm{c}_{mo}\right)\right)}},\\
\mathrm{c}_{mn}=\mathrm{a}\left[[h_{\mathrm{v}_m}^{t-1}||{r_{\mathrm{v}_{m}.{\mathrm{type}}}}]\mathrm{W}_q||[h_{\mathrm{v}_n}^{t-1}||r_{{\mathrm{v}_n}.\mathrm{type}}]\mathrm{W}_k\right]^{\mathrm{T}},
\end{eqnarray}
% \end{small}
where $\mathrm{c}_{mn}$ is the attentive coefficient score between $\mathrm{v}_{m}$ and $\mathrm{v}_{n}$, $[\cdot||\cdot]$ is the concatenating operation, $r_{\mathrm{v}_{m}.\mathrm{type}}\in\mathbb{R}^{1\times4}$ is a one-hot vector denoting the node type of $\mathrm{v}_{m}$, and $\mathrm{a}\in\mathbb{R}^{1\times 2d_{head}}$, $\mathrm{W}_{q}\in\mathbb{R}^{(d_{head}+4)\times d_{head}}$, $\mathrm{W}_{k}\in\mathbb{R}^{(d_{head}+4)\times d_{head}}$ are trainable parameters. 

Once obtaining the importance of adjacent nodes, the graph attention block aggregates the weighted message by:
\begin{equation}
    h_{\mathrm{v}_m}^{t\cdot head}=\sigma\left(\sum_{\mathrm{v}_o\in\mathcal{N}_{\mathrm{v}_m}}{\alpha_{mn}h_{\mathrm{v}_{o}}^{t-1}\mathrm{W}_{v}}\right),
\end{equation}
where $\sigma$ is the ELU function, and $\mathrm{W}_{o}\in\mathbb{R}^{d_{head}\times d_{head}}$ is a trainable parameter. By concatenating weighted messages from all heads, the $t$-$th$ graph attention block can update the node representation to $h_{\mathrm{v}_m}^{t}$. 

\subsection{Answer Extraction Training and Inference. }
After graph modeling, nodes in the QuISG are then mapped back into the original token sequence. As answers are extracted from the dialogue, only nodes in key utterances are considered. We locate the dialogue word (scene) node $v_x$ with its corresponding token representation $H_C[utter_i[x]]$ in $\mathcal{C}$, we then update the token representation by $H_C[utter_i[x]]$ += $h_{\mathrm{v}_x}^{T}$. For the speaker token representation $H_C[$Ross$_{i}$, Gellar$_{i}]$ in key utterances, the mapped dialogue speaker node $\mathrm{v}_{s_i}$ updates it by $H_C[$Ross$_{i}$, Gellar$_{i}]$ += [$h_{\mathrm{v}_{s_i}}^{T}$, $h_{\mathrm{v}_{s_i}}^{T}$]. As a speaker name $s_i$ may appear several times, we repeat adding duplicated $h_{\mathrm{v}_{s_i}}^{T}$ to the corresponding token representations. We denote the updated $H_C$ as $H_{C}^{'}$. 

\subsubsection{Training. }
Given $H_{C}^{'}$, the model computes the start and the end distributions by:
\begin{eqnarray}
Y_{srt}=\mathrm{softmax}(\mathrm{w}_{srt}{H_{C}^{'}}^{\mathrm{T}}), \\
Y_{end}=\mathrm{softmax}(\mathrm{w}_{end}{H_{C}^{'}}^{\mathrm{T}}),
\end{eqnarray}
where $\mathrm{w}_{srt}\in\mathbb{R}^{1\times L_C}$, $\mathrm{w}_{end}\in\mathbb{R}^{1\times L_C}$ are trainable parameters. 
For the answer span $a$, we denote its start index and end index as $a_{st}$ and $a_{ed}$ respectively. Therefore, the answer extracting objective is:
\begin{equation}
    \mathcal{J}_{ax}=-\mathrm{log}\left(Y_{srt}(a_{st})\right) - \mathrm{log}\left(Y_{end}(a_{ed})\right). 
\end{equation}

If there are questions without any answers, another header is applied to predict whether a question is answerable. The header computes the probability by $p_{na}=\mathrm{sigmoid}(\mathrm{Linear}(H_{C}^{'}[\mathrm{CLS}]))$. By annotating every question with a label $q\in\{0,1\}$ to indicate answerability, another objective is added:
\begin{equation}
    \mathcal{J}_{na}=-[(1-q)\mathrm{log}(1-p_{na})+q\mathrm{log}(p_{na})]. 
\end{equation}

In this way, the overall training objective is $\mathcal{J}=\mathcal{J}_{ax}+0.5*\mathcal{J}_{na}$. 

\subsubsection{Inference. }
Following \citet{SelfSuper}, we extract the answer span by performing beam search with the size of 5. We constrain the answer span in one utterance to avoid answers across utterances. To further emphasize the importance of key utterances, we construct a scaling vector $S\in\mathbb{R}^{1\times L_C}$, where the token belonging to key utterances is kept with 1 and the token out of key utterances is assigned with a scale factor $0\leq f\leq1$.  The scaling vector is multiplied on $Y_{srt}$ and $Y_{end}$ before softmax, and we then perform the ranking on the possibilities of answers for inference. 

\section{Experimental Settings}
\paragraph{Datasets.}
Following \citet{SelfSuper}, we conduct experiments on \textbf{FriendsQA} \citep{FriendsQA} and \textbf{Molweni} \citep{Molweni}. As our work does not focus on unanswerable questions, we construct an answerable version of Molweni (\textbf{Molweni-A}) by removing all unanswerable questions. FriendsQA is a open-domain question answering dataset collected from the TV-series \textit{Friends}. It contains 977/122/123 (train/dev/test) dialogues and 8,535/1,010/1,065 questions. As we recognize person names in questions, we find about 76\%/76\%/75\% of questions contain person names in FriendsQA. Molweni is another related dataset whose topic is about Ubuntu. It contains 8,771/883/100 dialogues and 24,682/2,513/2,871 questions, in which about 14\% questions are unanswerable. Dialogues in Mowelni is much shorter than those in FriendsQA and contain no scene descriptions. Speaker names in Molweni are meaningless user ids (e.g., ``nbx909''). Furthermore, questions containing user ids in Molweni, whose proportion is about 47\%/49\%/48\%, are much less than those in FriendsQA. In Molweni-A, there are 20,873/2,346/2,560 questions. 

\paragraph{Compared Methods.}
We compare our method with existing methods in DRC. \textbf{ULM+UOP} \citep{ULMUOP} adapt token MLM, utterance MLM, and utterance order prediction to pretrain BERT, and finetune it in the multi-task setting. \textbf{KnowledgeGraph} \citep{KnowledgeGraph} introduce and structurally model additional knowledge about speakers' co-reference and relations between each other from other related datasets \citep{RelationExtraction}. \textbf{DADGraph} \citep{DADGraph} is another graph-based method that introduce external knowledge about the discourse structure of dialogues. \textbf{ELECTRA} \citep{ELECTRA} is the vanilla pretrained model used as the backbone of our method. \textbf{SelfSuper} \citep{SelfSuper} is the SOTA method. It design two self-supervised tasks to capture speaker information and reduce noise in the dialogue. 

\paragraph{Implementation.}
Our medel is implemented based on \texttt{ELECTRA-large-discriminator} from \textit{Transformers}. For key utterances extraction, the size of the sliding window (i.e., $m$) is set to 2 and top-3 units are considered. Other hyper-parameters are the same as those in the question answering training. For question answering, we search the size of word node window (i.e., $k_{w}$) in 1, 2, 3, and the number of attention heads in 1, 2, 4. We set the number of GAT layers to 5 for FriendsQA and 3 for Molweni; $f$ is set to 0.5 for FriendsQA and 0.9 for Molweni. For other hyper-parameters, we follow \citet{SelfSuper}. We use Exact Matching (EM) score and F1 score as the metrics. 

\section{Results and Discussion}

\subsection{Main Results}

\begin{table}[]
\centering
\scalebox{0.9}{
\begin{tabular}{@{}clcc@{}}
\toprule
&\multicolumn{1}{l|}{Model}          & \multicolumn{1}{c}{EM}   & \multicolumn{1}{c}{F1}             \\ \midrule\midrule
\multirow{3}{*}{\rotatebox[origin=c]{90}{\begin{tabular}[c]{@{}c@{}} \scriptsize{BERT}\\\scriptsize{based}\\\end{tabular}}} & \multicolumn{1}{l|}{ULM+UOP \citep{ULMUOP}}        & \multicolumn{1}{c}{46.80}           & \multicolumn{1}{c}{63.10}         \\
&\multicolumn{1}{l|}{KnowledgeGraph \citep{KnowledgeGraph}} & \multicolumn{1}{c}{46.40}               & \multicolumn{1}{c}{64.30}              \\
&\multicolumn{1}{l|}{SelfSuper$_{\textrm{BERT}}$ \citep{SelfSuper}} & \multicolumn{1}{c}{46.90}           & \multicolumn{1}{c}{63.90} \\ 
\midrule
\multirow{3}{*}{\rotatebox[origin=c]{90}{\begin{tabular}[c]{@{}c@{}} \scriptsize{ELECTRA}\\\scriptsize{based}\\\end{tabular}}} & \multicolumn{1}{l|}{Our Reimpl. ELECTRA}        & \multicolumn{1}{c}{54.62}           & \multicolumn{1}{c}{71.29}          \\
&\multicolumn{1}{l|}{SelfSuper \citep{SelfSuper}} & \multicolumn{1}{c}{55.80}           & \multicolumn{1}{c}{72.30}           \\ 
% \midrule
\cmidrule(lr){2-4}
&\multicolumn{1}{l|}{Ours}           & \multicolumn{1}{c}{\textbf{57.79}$^{*}$} & \multicolumn{1}{c}{\textbf{75.22}$^{*}$} \\ \bottomrule
\end{tabular}
}
\caption{Results on FriendsQA. $^{*}$ denotes significance against SelfSuper with paired t-test. }
\label{tab1}
\end{table}

\begin{table}[]
\centering
\scalebox{0.9}{
\begin{tabular}{@{}cl|cc@{}}
\toprule
&\multicolumn{1}{l|}{Model}          & \multicolumn{1}{c}{EM}   & \multicolumn{1}{c}{F1}           \\ \midrule\midrule
\multirow{2}{*}{\rotatebox[origin=c]{90}{\begin{tabular}[c]{@{}c@{}} \scriptsize{BERT}\\\scriptsize{based}\\\end{tabular}}} &\multicolumn{1}{l|}{DADGraph \citep{DADGraph}}                 & \multicolumn{1}{c}{46.50}          & \multicolumn{1}{c}{61.50}          \\
&\multicolumn{1}{l|}{SelfSuper$_{\textrm{BERT}}$ \citep{SelfSuper}} & \multicolumn{1}{c}{49.20}          & \multicolumn{1}{c}{64.00}          \\ \midrule
\multirow{3}{*}{\rotatebox[origin=c]{90}{\begin{tabular}[c]{@{}c@{}} \scriptsize{ELECTRA}\\\scriptsize{based}\\\end{tabular}}} &\multicolumn{1}{l|}{Our Reimpl. ELECTRA}                  & \multicolumn{1}{c}{57.85}          & \multicolumn{1}{c}{72.17}          \\
&\multicolumn{1}{l|}{SelfSuper \citep{SelfSuper}}           & \multicolumn{1}{c}{58.00}          & \multicolumn{1}{c}{\textbf{72.90}} \\ 
% \midrule
\cmidrule(lr){2-4}
&\multicolumn{1}{l|}{Ours}                     & \multicolumn{1}{c}{\textbf{59.32}$^{*}$} & \multicolumn{1}{c}{\textbf{72.86}} \\ \midrule
&\multicolumn{1}{l|}{Human performance}        & \multicolumn{1}{c}{64.30}          & \multicolumn{1}{c}{80.20}          \\ \bottomrule
\end{tabular}
}
\caption{Results on Molweni. }
\label{tab2}
\end{table}

Tab.~\ref{tab1} shows the results achieved by our method and other baselines on FriendsQA. Baselines listed in the first three rows are all based on BERT~\cite{BERT}. We can see that SelfSuper achieves better and competitive results compared with ULM+UOP and KnowledgeGraph, where ULM+UOP considers no speaker information and KnowledgeGraph introduces external knowledge. This indicates the effectiveness of the self-supervised tasks for speaker and key utterance modeling of SelfSuper. When it comes to ELECTRA, the performance reach a new elevated level, which shows that ELECTRA is more suitable for DRC. By comparing with SelfSuper based on ELECTRA, our method can achieve significantly better performance. 
This improvement shows the advantage of both the higher coverage of answer-contained utterances by our proposed key utterances extractor and better graph representations to consider the question and interlocutor scopes by QuISG. 

Results on Molweni are listed in Tab.~\ref{tab2}.
Our approach still gives new state-of-the-art, especially a significantly improvement on EM scores.
However, the absolute improvement is smaller compared to that on FriendsQA.
This is mainly for two reasons.
First, the baseline results are close to the human performance on Molweni, so the space for improvement is also smaller.
Second, Molweni consists of unanswerable questions, which are not the main focus of our work.
% We can see that SelfSuper shows excellent performance using BERT while the performance against ELECTRA is not that remarkable. 
% We think the reason may be that the results are near the human performance and hard questions are challenges for models to cope with. Under such a circumstance, our method achieves better EM score and competitive F1 score against SelfSuper. 
To see how the unanswerable questions affect the results, we further show the performance of our method and the ELECRTA-based baselines on Molweni-A in Tab.~\ref{tab3}, i.e., the subset of Molweni with only answerable questions. We observe that our method still achieves better EM score against SelfSuper and gains slightly better F1 score, which indicates that our method can better deal with questions with answers.
As for unanswerable questions, we believe that better performance can be achieved with related-techniques plugged in our method, which we leave to future work.

By comparing the performance gains of our method in FriendsQA and Molweni, we can observe that our method is more significant in FriendsQA. We think the reason may be that (1) our key utterance extractor can cover more answer-contained utterances in FriendsQA, as will be shown in Fig.~\ref{fig:key}; (2) questions with speaker mention show more frequently in FriendsQA than in Molweni, and therefore QuISG can help achieve better graph representations in FriendsQA. On all accounts, this further demonstrates that our method alleviates the problems that we focus on. 

\begin{table}[]
\centering
\scalebox{0.9}{
\begin{tabular}{@{}l|cc@{}}
\toprule
\multicolumn{1}{l|}{Model}          & \multicolumn{1}{c}{EM}   & \multicolumn{1}{c}{F1}      \\ \midrule\midrule
\multicolumn{1}{l|}{Our Reimpl. ELECTRA} & \multicolumn{1}{c}{61.02}   & \multicolumn{1}{c}{77.62}   \\
\multicolumn{1}{l|}{SelfSuper \citep{SelfSuper}} & \multicolumn{1}{c}{61.13}   & \multicolumn{1}{c}{78.30}   \\ \midrule
\multicolumn{1}{l|}{Ours}    & \multicolumn{1}{c}{\textbf{62.54}$^{*}$} & \multicolumn{1}{c}{\textbf{78.65}} \\ \bottomrule
\end{tabular}
}
\caption{Results on Molweni-A. }
\label{tab3}
\end{table}

\begin{table}[]
\centering
\scalebox{0.9}{
\begin{tabular}{@{}l|cccc@{}}
\toprule
\multirow{2}{*}{\ \ \ Model} 
                       & \multicolumn{2}{c|}{FriendsQA}     & \multicolumn{2}{c}{Molweni} \\ 
                       & \multicolumn{1}{c}{EM}    & \multicolumn{1}{c|}{F1}    & \multicolumn{1}{c}{EM}  & \multicolumn{1}{c}{F1}           \\ \midrule\midrule
\multicolumn{1}{l|}{full model}             & \multicolumn{1}{c}{\textbf{57.79}} & \multicolumn{1}{c|}{\textbf{75.22}} & \multicolumn{1}{c}{\textbf{59.32}}        & \multicolumn{1}{c}{\textbf{72.86}}        \\ \midrule
\multicolumn{1}{l|}{w/o NodeType}              & \multicolumn{1}{c}{56.79} & \multicolumn{1}{c|}{74.01} & \multicolumn{1}{c}{58.38}        & \multicolumn{1}{c}{72.75}        \\
\multicolumn{1}{l|}{w/o KeyUttExt}             & \multicolumn{1}{c}{55.87}     & \multicolumn{1}{c|}{72.30}     & \multicolumn{1}{c}{58.48}            & \multicolumn{1}{c}{72.10}            \\
\multicolumn{1}{l|}{w/o Q}                     & \multicolumn{1}{c}{56.37} & \multicolumn{1}{c|}{73.55} & \multicolumn{1}{c}{58.20}            & \multicolumn{1}{c}{72.52}            \\
\multicolumn{1}{l|}{w/o SpkScope}              & \multicolumn{1}{c}{57.29} & \multicolumn{1}{c|}{74.26} & \multicolumn{1}{c}{58.62}            & \multicolumn{1}{c}{72.29}            \\
\multicolumn{1}{l|}{w/o All}                   & \multicolumn{1}{c}{53.12}     & \multicolumn{1}{c|}{70.05}     & \multicolumn{1}{c}{56.32}            & \multicolumn{1}{c}{71.08}            \\ \bottomrule
\end{tabular}
}
\caption{Ablation Study.}
\label{tab4}
\end{table}

\subsection{Ablation Study}
To demonstrate the importance of our proposed modules, we adapt ablation study on both datasets. The results are shown in Tab.~\ref{tab4}. We study the effects of node type information (NodeType), key utterances extractor and its scaling factor on logits (KeyUttExt); question and question speaker nodes (Q); edges between dialogue word nodes and dialogue speaker nodes to model interlocutor scope (SpkScope). We further remove both KeyUttExt and QuISG, leading to fully connections between every two tokens in dialogues, and apply transformer layers to further process dialogues (w/o All).

By removing NodeType, the performance drops, which demonstrates minding different node behaviors can help better model graph representations. Our method w/o KeyUttExt decreases the performance, which demonstrates that key utterance extractor is a crucial module for our method to find more answer-contained utterances and guides our model to pay more attention to the key part in a dialogue for the question. As for the model w/o KeyUttExt shows more performance drop in FriendsQA, we think the reason may be that dialogues in FriendsQA are much longer than Molweni. Therefore, KeyUttExt can reduce more question unrelated parts of dialogues for further graph modeling in FriendsQA. Removing Q or SpkScope also shows performance decline, which indicates the importance of realizing question and interlocutor scope for DRC. The worse results of w/o Q indicates that questions and speakers mentioned in questions are more important to DRC due to the question-involved particularity of DRC. Replacing KeyUttExt and QuISG with transformer layers even performs worse than ELECTRA, which indicates that the further process of dialogues without speaker and question realized modeling can be redundant. 

\subsection{Accuracy of Utterance Extraction}

\begin{figure}
    \centering
    \scalebox{0.26}{\includegraphics{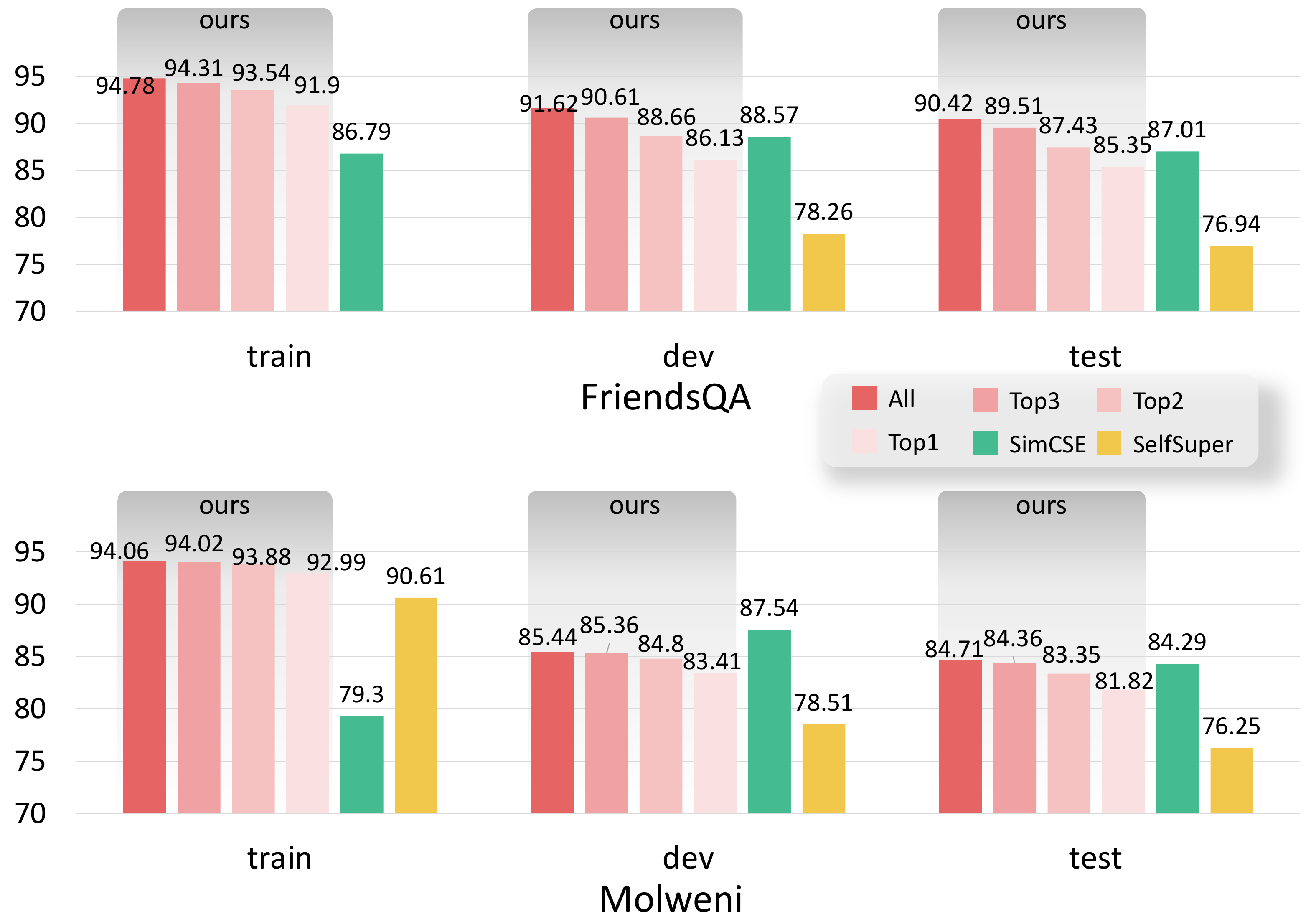}}
    \caption{Recall of answer coverage using different key utterance extracting methods. }
    \label{fig:key}
\end{figure}

As we claim that our method covers more answer-contained utterances compared with SelfSuper, in this section, we show the recall of answer-contained utterances by different methods. Besides our method and SelfSuper, we further consider retrieval methods appearing other reading comprehension tasks. As the similarity-based seeker is usually used, we apply the SOTA model SimCSE~\cite{SimCSE} to compute similarity score between utterances and the question. However, directly using top similar utterances produces a extremely low recall. Therefore, we also add utterances around every picked top utterance as key utterances like ours. We consider top 3 similar utterances and its historical 2 and future 2 utterances as key utterances. The results are illustrated in Fig.~\ref{fig:key}. As shown in Fig.~\ref{fig:key}, we choosing top 3 units for key utterances selection does not affect the recall a lot and can keep the average size of key utterance set to 4.13 for FriendsQA and 3.61 for Molweni. Compared with our method, SelfSuper achieves undesirable recall for answer-contained utterance extraction, which indicates the efficacy of our method. As for SimCSE, equipped with our enhancement, it can achieve competitive recall to ours. Especially on the dev and test sets of Molweni. However, the average size of key utterance set of SimCSE is 7.73, where the average length of dialogue in Molweni is 8.82. Additionally, SimCSE cannot judge whether a question is answerable and therefore extracts key utterances for every question, leading to a low recall for the training set in Molweni. 

%average length of dialogues in friendsqa is 21.92. 

To further demonstrate that our method is more suitable for DRC against SimCSE, we run a variant with key utterances sets extracted by SimCSE. The results are shown in Tab.~\ref{tab5}. Our method achieves better performance with high coverage of answer-contained utterances and fewer key utterances involved. 

\subsection{Improvement on Speaker-Contained Questions}

\begin{table}[]
\centering
\scalebox{0.9}{
\begin{tabular}{@{}l|cc|cc@{}}
\toprule
\multirow{2}{*}{\ \ Method} & \multicolumn{2}{c|}{FriendsQA} & \multicolumn{2}{c}{Molweni} \\ 
                        & \multicolumn{1}{c}{EM}            & \multicolumn{1}{c|}{F1}            & \multicolumn{1}{c}{EM}           & \multicolumn{1}{c}{F1}           \\ \hline\hline
\multicolumn{1}{l|}{ours}                & \multicolumn{1}{c|}{57.79}          & \multicolumn{1}{c|}{75.22}         & \multicolumn{1}{c|}{59.32}        & \multicolumn{1}{c}{72.86}        \\ \hline
\multicolumn{1}{l|}{SimCSE}              & \multicolumn{1}{c|}{57.19}              & \multicolumn{1}{c|}{74.36}             & \multicolumn{1}{c|}{57.30}            & \multicolumn{1}{c}{72.09}            \\ \bottomrule
\end{tabular}
}
\caption{Results of our method and the variant with SimCSE searching for key utterances. }
\label{tab5}
\end{table}

\begin{figure}
    \centering
    \scalebox{0.24}{\includegraphics{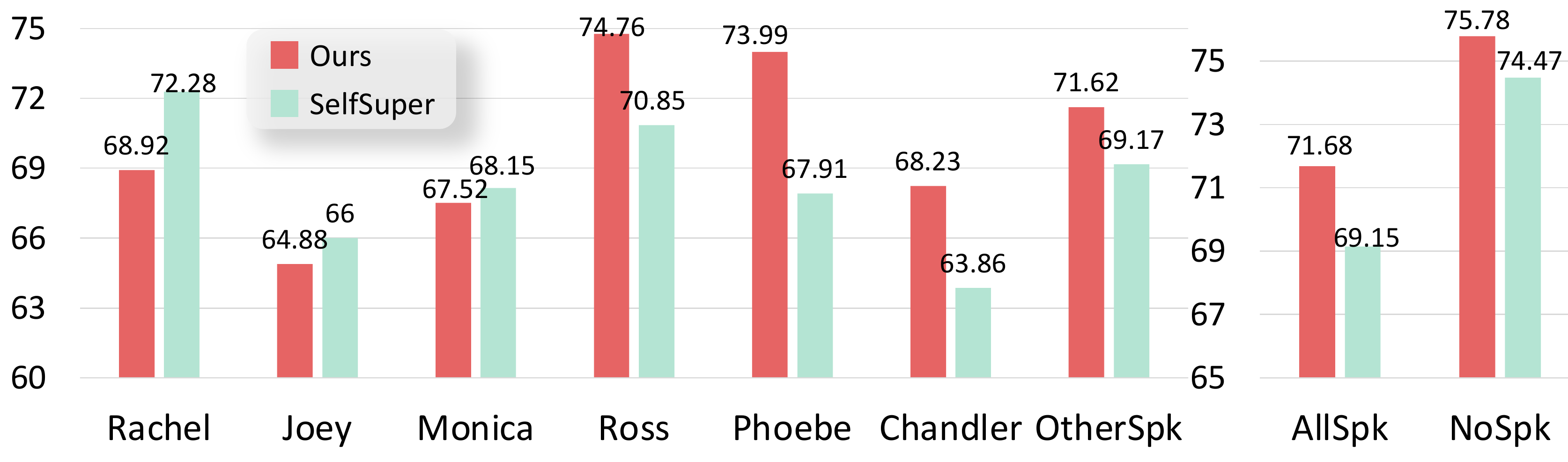}}
    \caption{F1 scores of answers to the questions with or without speaker names in the dev set of FriendsQA. }
    \label{spk}
\end{figure}

As QuISG focuses on the question speaker information and dialogue interlocutor scope modeling, whether it can help answer questions mentioning speaker names is crucial to verify. We illustrate the F1 scores of questions containing different speakers in FriendsQA and questions with or without mentioning speakers in Fig.~\ref{spk}. We can see that SelfSuper outperforms our method only on ``\texttt{Rachel}'' and is slightly better on ``\texttt{Joey}'' and ``\texttt{Monica}''. Our method can outperform SelfSuper by a great margin on ``\texttt{Ross}'', ``\texttt{Phoebe}'', ``\texttt{Chandler}'' and other casts. Specifically, our method can improve the F1 score of questions mentioning speaker by a wider margin compared to questions without speakers. This indicates that our speaker modeling is benefit from our proposed method. 

\subsection{Case Study}

As shown in the very beginning of the paper, Fig.~\ref{fig:case} provides two cases that SelfSuper fails. On the contrary, attributing to our proposed key utterances extractor and QuISG, our method can answer the two questions correctly. 

\section{Conclusion}

To cover more answer-contained utterances and make the model realize speaker information in the question and interlocutor scopes in the dialogue for DRC, we propose a new pipeline method. The method firstly adapts a new key utterances extractor with contiguous utterances as a unit for prediction. Based on utterances in the extracted units, a Question-Interlocutor Scope Realized Graph (QuISG) is constructed. QuISG sets question-mentioning speakers as question speaker nodes and connects the speaker node in the dialogue with words from its scope. Our proposed method achieves decent performance on related benchmarks. 
% Use \bibliography{yourbibfile} instead or the References section will not appear in your paper
\bibliography{mybib}

\end{document}